\documentclass[runningheads]{llncs}
\usepackage{graphicx}

\usepackage{tikz}
\usepackage{comment}
\usepackage{amsmath,amssymb} 
\usepackage{color}

\usepackage[accsupp]{axessibility}  

\usepackage{caption}
\usepackage{subcaption}
\usepackage{bbold}
\usepackage{colortbl}
\usepackage{float}
\usepackage{bbm}
\usepackage{upgreek}
\usepackage{comment}
\usepackage{bm}
\usepackage{booktabs}
\usepackage[pagebackref,breaklinks,colorlinks]{hyperref}
\usepackage[capitalize]{cleveref}
\crefname{section}{Sec.}{Secs.}
\Crefname{section}{Section}{Sections}
\Crefname{table}{Table}{Tables}
\crefname{table}{Tab.}{Tabs.}

\begin{document}
\pagestyle{headings}
\mainmatter
\def\ECCVSubNumber{69}  

\definecolor{mapillarygreen}{RGB}{38,235,179}

\definecolor{blue1}{RGB}{86,174,139}
\definecolor{tabgray}{gray}{.9}

\newcommand{\MS}[1]{\textcolor{red}{{\bf #1}}}
\newcommand{\ms}[1]{\textcolor{red}{#1}}
\newcommand{\ZD}[1]{{\color{blue}{\bf ZD: #1}}}
\newcommand{\zd}[1]{{\color{blue}{#1}}}
\newcommand{\YG}[1]{{\color{orange}{\bf JNQ: #1}}}
\newcommand{\yg}[1]{{\color{orange}{#1}}}
\newcommand{\LML}[1]{{\color{green}{\bf LML: #1}}}
\newcommand{\lml}[1]{{\color{green}{#1}}}
\newcommand{\LZW}[1]{{\color{cyan}{\bf LZW: #1}}}
\newcommand{\lzw}[1]{{\color{cyan}{#1}}}

\newcommand{\bX}{\mathcal{X}}
\newcommand{\btX}{\tilde{\mathcal{X}}}
\newcommand{\bx}{\mathbf{x}}
\newcommand{\bbx}{\bar{\mathbf{x}}}
\newcommand{\bY}{\mathcal{Y}}
\newcommand{\by}{\mathbf{y}}
\newcommand{\bby}{\bar{\mathbf{y}}}
\newcommand{\bfx}{\mathbf{f}_x}
\newcommand{\bfy}{\mathbf{f}_y}
\newcommand{\bthx}{\theta_x}
\newcommand{\bthy}{\theta_y}
\newcommand{\bthxt}{\tilde{\theta}_x}
\newcommand{\bthyt}{\tilde{\theta}_y}
\newcommand{\bK}{\mathbf{K}}
\newcommand{\bT}{\mathcal{T}}
\newcommand{\bt}{\mathbf{t}}
\newcommand{\bS}{\mathcal{S}}
\newcommand{\bbS}{\mathcal{\bar{S}}}
\newcommand{\bP}{\mathcal{P}}
\newcommand{\bbP}{\mathcal{\bar{P}}}
\newcommand{\bM}{\mathcal{M}}
\newcommand{\bQ}{\mathcal{Q}}
\newcommand{\bbM}{\mathcal{\bar{M}}}
\newcommand{\bA}{\mathcal{A}}
\newcommand{\bH}{\mathbf{H}}
\newcommand{\bR}{\mathbf{R}}
\newcommand{\bW}{\mathbf{W}}
\newcommand{\bU}{\mathbf{U}}
\newcommand{\bI}{\mathbf{I}}
\newcommand{\bV}{\mathbf{V}}
\newcommand{\ba}{\mathbf{a}}
\newcommand{\bb}{\mathbf{b}}
\newcommand{\bg}{\mathbf{g}}
\newcommand{\bo}{\mathbf{o}}
\newcommand{\bho}{\mathbf{\hat{o}}}

\newcommand{\hR}{\hat{R}}
\newcommand{\hht}{\hat{\textbf{t}}}
\newcommand{\gR}{R_{gt}}
\newcommand{\gt}{\textbf{t}_{gt}}
\newcommand{\norm}[1]{\left\lVert#1\right\rVert}
\newcommand{\argmin}{\mathop{\mathrm{argmin}}}

\newcommand\largeheight{0.09\columnwidth}

\title{Learning-based Point Cloud Registration for 6D Object Pose Estimation in the Real World} 

\titlerunning{Match Normalization}
%

\author{Zheng Dang\inst{1}\index{Dang, Zheng} \and
Lizhou Wang\inst{2} \and
Yu Guo\inst{2} \and
Mathieu Salzmann\inst{1,3}}

%
\authorrunning{Z. Dang et al.}
%
\institute{CVLab, EPFL, Switzerland \\
\email{\{zheng.dang, mathieu.salzmann\}@epfl.ch}\\
Xi'an Jiaotong University, Shaanxi, China\\
\email{dzyxwanglizhou@stu.xjtu.edu.cn, yu.guo@xjtu.edu.cn}\\
Clearspace, Switzerland}
\maketitle
\begin{abstract}
In this work, we tackle the task of estimating the 6D pose of an object from point cloud data. While recent learning-based approaches to addressing this task have shown great success on synthetic datasets, we have observed them to fail in the presence of real-world data. We thus analyze the causes of these failures, which we trace back to the difference between the feature distributions of the source and target point clouds, and the sensitivity of the widely-used SVD-based loss function to the range of rotation between the two point clouds. We address the first challenge by introducing a new normalization strategy, Match Normalization, and the second via the use of a loss function based on the negative log likelihood of point correspondences. Our two contributions are general and can be applied to many existing learning-based 3D object registration frameworks, which we illustrate by implementing them in two of them, DCP and IDAM. Our experiments on the real-scene TUD-L~\cite{Hodan18}, LINEMOD~\cite{Hinterstoisser12} and Occluded-LINEMOD~\cite{Brachmann14} datasets evidence the benefits of our strategies. They allow for the first time learning-based 3D object registration methods to achieve meaningful results on real-world data. We therefore expect them to be key to the future development of point cloud registration methods. Our source code can be found at \href{https://github.com/Dangzheng/MatchNorm}{https://github.com/Dangzheng/MatchNorm}.

\end{abstract}

\section{Introduction}
Estimating the 6D pose, i.e., 3D rotation and 3D translation, of an object has many applications in various domains, such as robotics grasping, simultaneous localization and mapping (SLAM), and augmented reality. In this context, great progress has been made by learning-based methods operating on RGB(D) images~\cite{Labbe20,Rad17,Peng19,Tremblay18,Park19,Zakharov19,Wang19a,Li18a,Wang19,Martin18}. In particular, these methods achieve impressive results on real-world images.


\begin{figure}[!ht]
    \centering
    \includegraphics[width=\textwidth]{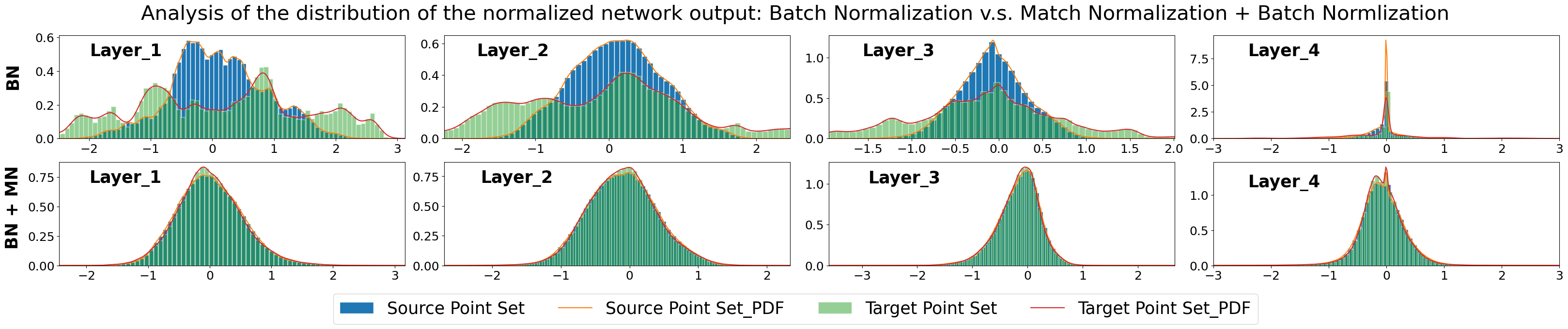}
    \caption{\label{fig:distribution} {\bf Feature distributions at different network layers with real-world data.} {\bf Top:} With Batch Normalization, the distributions of the features extracted from the model (source) and input (target) point clouds differ significantly. {\bf Bottom:} Our Match Normalization strategy makes these distributions much more similar.}

\end{figure}
In parallel to this line of research, and thanks to the development of point cloud processing networks~\cite{Qi17,Qi17a,Atzmon18,Wang18b,Li18c,Su18,Zaheer17}, several learning-based 3D object registration algorithms~\cite{Aoki19,Wang19e,Wang19f,Yew20,Yuan20} have emerged to estimate 6D object poses from 3D measurements only, such as those obtained with a LiDAR. Since they focus solely on 3D information, discarding any RGB appearance, these methods have demonstrated excellent generalization to previously unseen objects. However, in contrast with \emph{scene-level} registration methods~\cite{Choy20,Choy19a,Deng18,Deng18a,Huang21,Cao21}, these \emph{object-level} learning-based techniques are typically evaluated only on synthetic datasets, and virtually never on real-world data, such as the TUD-L~\cite{Hodan18}, LineMod~\cite{Hinterstoisser12} and LineMod-Occluded~\cite{Brachmann14} datasets.

In our experiments with the state-of-the-art learning-based object-level registration frameworks, we observed them to struggle with the following challenges. First, in contrast with synthetic datasets where all objects have been normalized to a common scale, the size of different objects in the real world varies widely. The fact that the sensor depicts only an unknown portion the object precludes a simple re-scaling of the target point cloud. In synthetic data, the target point cloud is typically sampled from the normalized model, thus ignoring this difficulty. Second, while synthetic datasets typically limit the rotation between the source and target point clouds in the $45^\circ$ range, real-world sensors may observe the target object from any viewpoint, covering the full rotation range.

As shown in the top row of Figure~\ref{fig:distribution}, the first above-mentioned challenge translates to a significant gap between the feature distributions of the source and target point clouds in the inner layers of the network.
The greater the difference between the two distributions, the fewer correct inlier matches will be found, which then yields a decrease in performance. 
To address this, we propose a new normalization method, which we refer to as Match Normalization. Match Normalization exploits an instance-level scale parameter in each layer of the feature extraction network. This parameter is shared by the source and target point clouds, thus providing robustness to partial observations and outliers. 
This makes the distributions of the two point clouds more concentrated and similar, as shown in the bottom row of Figure~\ref{fig:distribution}, and yields increase the number of inlier matches, as evidenced by our experiments.


Furthermore, we observed the second above-mentioned challenge to cause instabilities in the network convergence when relying on the widely-used SVD-based loss function~\cite{Wang19e,Wang19f,Yew20} for training. To address this, we exploit a simple negative log-likelihood (NLL) loss function, which we show to improve convergence and lead to better object-level pose estimation accuracy.

Altogether, our contributions have the following advantages: (i) The proposed Match Normalization is applicable to many point cloud registration network architectures; (ii) Both contributions only involve small changes to the network and yet substantially improve its performance on real object-level pose estimation datasets; (iii) They allow for the first time a learning-based point cloud registration method to achieve meaningful results on real-world 6D object pose estimation datasets, such as the TUD-L~\cite{Hodan18}, LINEMOD~\cite{Hinterstoisser12} and Occluded-LINEMOD~\cite{Brachmann14} datasets. We release our code to facilitate reproducibility and future research.


\section{Related Work}

\textbf{Traditional Point Cloud Registration Methods.} The Iterative Closest Point (ICP)~\cite{Besl92a} is the best-known local registration methods.Several variants, such as Generalized-ICP~\cite{Segal09} and Sparse ICP~\cite{Bouaziz13}, have been proposed to improve robustness to noise and mismatches, and we refer the reader to~\cite{Pomerleau15,Rusinkiewicz01} for a complete review of ICP-based strategies. The main drawback of these methods is their requirement for a reasonable initialization to converge to a good solution. Only relatively recently has this weakness been addressed by the globally-optimal registration method Go-ICP~\cite{Yang15}. In essence, this approach follows a branch-and-bound strategy to search the entire 3D motion space $SE(3)$. A similar strategy is employed by the sampling-based global registration algorithm Super4PCS~\cite{Mellado14}.
While globally optimal, Go-ICP come at a much higher computational cost than vanilla ICP. This was, to some degree, addressed by the Fast Global Registration (FGR) algorithm~\cite{Zhou16}, which leverages a local refinement strategy to speed up computation. While effective, FGR still suffers from the presence of noise and outliers in the point sets, particularly because, as vanilla ICP, it relies on 3D point-to-point distance to establish correspondences. In principle, this can be addressed by designing point descriptors that can be more robustly matched. For example, \cite{Vidal18} relies on generating pose hypotheses via feature matching, followed by a RANSAC-inspired method to choose the candidate pose with the largest number of support matches. Similarly, TEASER~\cite{Yang19} and its improved version TEASER++~\cite{Yang20} take putative correspondences obtained via feature matching as input and remove the outlier ones by an adaptive voting scheme. In addition to the above, there are many algorithms~\cite{Aiger08,Mellado14,Raposo17,Mohamad15,Drost10,Maron16,Rosen19,Jzatt20,Johnson99,Rusu08,Rusu09,Yang19,Le19,Agamennoni16,Hinzmann16,Hahnel02,Fitzgibbon03,Bronstein09,Bronstein08,Gelfand05,Litany12} that have contributed to this direction.

\textbf{Learning-based Object Point Cloud Registration Methods.}
A key requirement to enable end-to-end learning-based registration was the design of deep networks acting on unstructured sets. Deep sets~\cite{Zaheer17} and PointNet~\cite{Qi17} constitute the pioneering works in this direction. 
In particular, PointNetLK~\cite{Aoki19} combines the PointNet backbone with the traditional, iterative Lucas-Kanade (LK) algorithm~\cite{Lucas81} so as to form an end-to-end registration network; DCP~\cite{Wang19e} exploits DGCNN~\cite{Wang18b} backbones followed by Transformers~\cite{Vaswani17} to establish 3D-3D correspondences, which are then passed through an SVD layer to obtain the final rigid transformation. While effective, PointNetLK and DCP cannot tackle the partial-to-partial registration scenario. That is, they assume that both point sets are fully observed, during both training and test time. This was addressed by PRNet~\cite{Wang19f} via a deep network designed to extract keypoints from each input set and match these keypoints. This network is then applied in an iterative manner, so as to increasingly refine the resulting transformation. IDAM~\cite{Li20} builds on the same idea as PRNet, using a two-stage pipeline and a hybrid point elimination strategy to select keypoints.
By contrast, RPM-Net~\cite{Yew20} builds on DCP and adopts a different strategy, replacing the softmax layer with an optimal transport one so as to handle outliers. Nevertheless, as PRNet, RPM-Net relies on an iterative strategy to refine the computed transformation. DeepGMR~\cite{Yuan20} leverages mixtures of Gaussians and formulates registration as the minimization of the KL-divergence between two probability distributions to handle outliers.
In any event, the methods discussed above were designed to handle point-clouds in full 3D, and were thus neither demonstrated for registration from 2.5D measurements, nor evaluated on real scene datasets, such as TUD-L, LINEMOD and Occluded-LINEMOD. In this work, we identify and solve the issues that prevent the existing learning-based methods from working on real-world data, and, as a result, develop the first learning-based point cloud registration method able to get reasonable result on the real-world 6D pose estimation datasets.

\section{Methodology}

\subsection{Problem Formulation}
Let us now introduce our approach to object-level 3D registration.
We consider the problem of partial-to-whole registration between two point clouds $\bX = \{\bx_1, \cdots, \bx_M\}$ and $\bY = \{\by_1, \cdots, \by_N\}$, which are two sets of 3D points sampled from the same object surface, with $\bx_i, \by_j \in \mathbb{R}^3$. We typically refer to $\bX$ as the source point set, which represents the whole object, and to $\bY$ as the target point set, which only contains a partial view of the object. We obtain the source point set $\bX$ by uniform sampling from the mesh model, and the target one $\bY$ from a depth map $I_{depth}$ acquired by a depth sensor, assuming known camera intrinsic parameters.  Our goal is to estimate the rotation matrix $\bR\in SO(3)$ and translation vector $\bt \in \mathbb{R}^{3}$ that align $\bX$ and $\bY$. The transformation $\bR, \bt$ can be estimated by solving
\begin{equation}
    \min_{\bR,\bt} = \sum_{\bx \in \bX_s} \norm{\bR\bx + \bt - \bQ(\bx)}^{2}_{2},
\label{eq:trans}
\end{equation}
where $\bQ:\bX_s\rightarrow\bY_s$ denotes the function that returns the best matches from set $\bX_s$ to set $\bY_s$, with $\bX_s$ and $\bY_s$ the selected inlier point sets.
Given the matches, Eq.~\ref{eq:trans} can be solved via SVD~\cite{Besl92a,Gower75}. The challenging task therefore is to estimate the matching function $\bQ$, with only $\bX$ and $\bY$ as input.

\subsection{Method Overview}

Most learning-based 3D registration methods rely on an architecture composed of two modules: the feature extraction module and the point matching module. The feature extraction module takes the two point sets as input, and outputs a feature vector $\bfx^{(i)}$, resp. $\bfy^{(j)}$, for either each point~\cite{Wang19e,Yew20} in $\bX$, resp. $\bY$, or each keypoint~\cite{Wang19f,Li20} extracted from $\bX$, resp. $\bY$. 

Given these feature vectors, a score map $\bS \in \mathbb{R}^{M\times N}$ is formed by computing the similarity between each source-target pair of descriptors. That is, the $(i,j)$-th element of $\bS$ is computed as
\begin{equation}
    \bS_{i,j} = <\bfx^{(i)}, \bfy^{(j)}>, \;\;\forall (i, j) \in [1,M]\times [1,N]\;,
    \label{eq:score}
\end{equation}
where $<\cdot, \cdot>$ is the inner product, and $\bfx^{(i)}, \bfy^{(j)} \in \mathbb{R}^P$. This matrix is then given to the point matching module, whose goal is to find the correct inlier matches between the two point sets while rejecting the outliers.

The parameters of the network are typically trained by minimizing the difference between the ground-truth rotation and translation and those obtained by solving Eq.~\ref{eq:trans}. 
Because, given predicted matches, the solution to Eq.~\ref{eq:trans} can be obtained by SVD, training can be achieved by backpropagating the gradient through the SVD operator, following the derivations in~\cite{Ionescu15}.


In the remainder of this section, we first introduce our approach to robustifying the feature extraction process, and then discuss the loss function we use to stabilize the training process in the presence of a full rotation range. Finally, we provide the details of the two models, DCPv2~\cite{Wang19e} and IDAM~\cite{Li20}, in which we implemented our strategies.
\begin{figure}[!t]
    \centering
    \includegraphics[width=\textwidth]{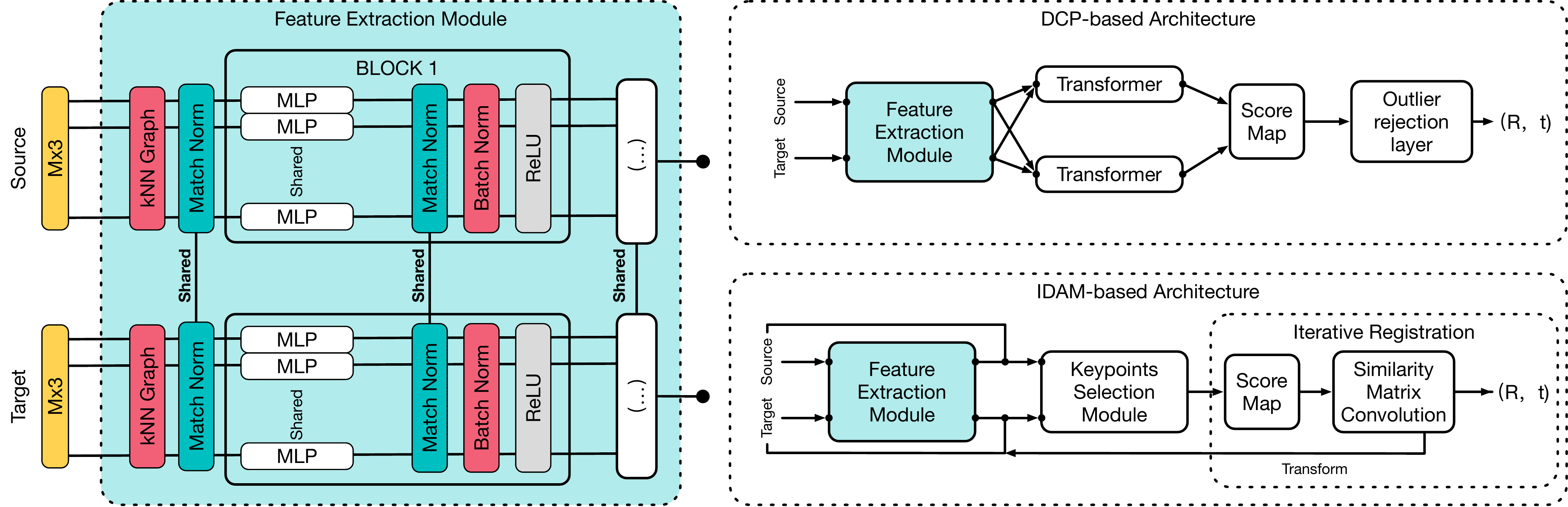}
    \caption{\label{fig:arch} {\bf Architecture of our DCP-based and IDAM-based frameworks.} We integrate Match Normalization in every block of the feature extraction module, sharing the scale parameter between the corresponding source and target point sets. The feature extraction module is central to the success of 3D registration architectures, such as DCP and IDAM. Its outputs greatly affect the subsequent score maps, and thus the construction of matches and ultimately the pose estimates. Our Match Normalization strategy yields robust features in the presence of real-world data.}

\end{figure}

\subsection{Match Normalization for Robust Feature Extraction}

Most learning-based 3D registration methods~\cite{Wang19e,Yew20,Wang19f,Yuan20,Li20} build on the PointNet~\cite{Qi17} architecture for the feature extraction module. Specifically, they use either convolutional layers or MLPs operating on a graph or the raw point set, with the output of each layer being normalized by Batch Normalization followed by ReLU. The normalized features of each layer then go through an additional subnetwork, which aggregates them into global features that are concatenated to the point-wise ones.

In this process, Batch Normalization aims to standardize the feature distributions to speed up training. Batch normalization assumes that every sample follows the same global statistics. In synthetic datasets, where all point clouds have been normalized to a common size, 
this assumption is typically satisfied. However, real-world data obtained by capturing objects with a depth sensor often includes objects of highly diverse sizes and only depicts unknown portions of these objects.
Therefore, the data does not meet the Batch Normalization assumption, and using the same normalization values for all samples within a mini-batch yields a large gap in the distributions of the features extracted by the network. This is illustrated in the top row of Figure~\ref{fig:distribution}, where we show the histogram and corresponding probability density function of the output of each layer of a network trained with Batch Normalization. Specifically, for each layer, we show a histogram encompassing all the feature channels for all the points in the source point cloud, and a similar histogram for the corresponding target point cloud.
Additional examples are provided in the supplementary material.
These plots clearly highlights the differences between the distributions of the corresponding source and target point clouds. In practice, these differences then affect the number of matched points, ultimately leading to low registration accuracy.


To overcome this, we introduce Match Normalization. The central idea between Match Normalization is to force the two point sets to have similar distributions. Specifically, we achieve this by centering the source and target point sets separately but by scaling them with the same parameter. The resulting normalized features then satisfy the Batch Normalization assumption, and, as shown in Figure~\ref{fig:arch}, we then feed them into a Batch Normalization layer to retain the benefit of fast training.

Formally, Match Normalization can be expressed as follows. For a layer with $C$ output channels, let $\bo_{x}^{} \in \mathbb{R}^{C\times M}$ and $\bo_{y} \in \mathbb{R}^{C\times N}$ be the features obtained by processing $\bX$ and $\bY$, respectively. We then normalize the features for each point $i$ as
\begin{equation}
\begin{aligned}
\bho_{x}^{(i)} = \frac{1}{\beta}(\bo_{x}^{(i)} - \mu_{x}), \quad\quad \bho_{y}^{(i)} = \frac{1}{\beta}(\bo_{y}^{(i)} - \mu_{y}), \quad
\end{aligned}
\end{equation}
where $\mu_x = \frac{1}{M}\sum^{M}_{i=1}\bo_x^{(i)}$, and similarly for $\mu_y$, are calculated separately for $\bo_{x}$ and $\bo_{y}$, but the scale 
\begin{equation}
\begin{aligned}
\beta = \max_{(i,j)=(1,1)}^{(M,C)} |\bo_{x}^{(i,j)}|\;,
\end{aligned}
\end{equation}
where $\bo_{x}^{(i,j)}$ denotes the $j$-th feature of the $i$-th point,
is shared by the corresponding source and target point sets. This scale parameter is computed from the source features, which are not subject to partial observations as the source points are sampled from the object model.

One advantage of using the same scale parameter for both point sets is robustness to partial observations and to outliers. Indeed, if the target point cloud had its own scaling parameter, the presence of partial observations, respectively outlier measurements, in the target point cloud might lead to stretching, respectively squeezing, it. By contrast, the source point cloud is complete and does not contain outliers. We thus leverage the intuition that the source and target point sets should be geometrically similar, and use the same scaling parameters in the Match Normalization process.


\subsection{NLL Loss Function for Stable Training}

\begin{table*}[!t]
    \centering
    {\scriptsize
    \setlength{\tabcolsep}{1.2mm}{
    \begin{tabular}{l|cccccc|cccccc}
        \toprule
        & \multicolumn{6}{c|}{Rotation mAP} & \multicolumn{6}{c}{Translation mAP} \\
        Method & $5^{\circ}$ & $10^{\circ}$ & $15^{\circ}$ & $20^{\circ}$ & $25^{\circ}$ & $30^{\circ}$& $0.001$ & $0.005$ & $0.01$ & $0.05$ & $0.1$ & $0.5$ \\
        \midrule
        DCP(v2)+SVD ($45^\circ$)   & 0.36 & 0.75 & 0.90 & 0.96 & 0.98 & 0.99 & 0.51 & 0.92 & 0.98 & 1.00 & 1.00 & 1.00 \\ 
        DCP(v2)+NLL ($45^\circ$)   & 0.72 & 0.97 & 1.00 & 1.00 & 1.00 & 1.00 & 0.54 & 0.99 & 1.00 & 1.00 & 1.00 & 1.00 \\ 
        DCP(v2)+SVD (Full) & 0.00 & 0.01 & 0.02 & 0.03 & 0.04 & 0.05 & 0.04 & 0.23 & 0.42 & 0.96 & 1.00 & 1.00 \\ 
        DCP(v2)+NLL (Full) & 0.35 & 0.67 & 0.86 & 0.94 & 0.97 & 0.98 & 0.19 & 0.57 & 0.85 & 1.00 & 1.00 & 1.00 \\ 

        \bottomrule
    \end{tabular}
    }}
    \caption{{\bf Influence of the loss function.} The models are evaluated on the partial-to-partial registration task on ModelNet40 (clean) as in~\cite{Wang19f,Yew20,Li20}, with either a $45^\circ$ rotation range, or a full one. For a given rotation range, the NLL loss yields better results than the SVD-based one. In the full rotation range scenario, the SVD-based loss fails completely, while the NLL loss still yields reasonably accurate pose estimates.
    }
    \label{tab:loss_func}
\end{table*}
In the commonly-used synthetic setting, the relative rotation between the two point clouds is limited to the $[0^{\circ}, 45^{\circ}]$ range. By contrast, in real data, the objects' pose may cover the full rotation range. To mimic this, we use the synthetic ModelNet40 dataset and modify the augmentation so as to generate samples in the full rotation range. As shown in Table~\ref{tab:loss_func}, our DCPv2 baseline, although effective for a limited rotation range, fails in the full range setting.
Via a detailed analysis of the training behavior, we traced the reason for this failure back to the choice of loss function. Specifically, the use of an SVD-based loss function yields instabilities in the gradient computation. This is due to the fact that, as can be seen from the mathematical expression of the SVD derivatives in~\cite{Ionescu15}, when two singular values are close to each other in magnitude, the derivatives explode.

To cope with this problem, inspired by~\cite{Sarlin19}, we propose to use the negative log likelihood loss to impose a direct supervision on the score map. To this end, let $\bM \in \{0, 1\}^{M \times N}$ be the matrix of ground-truth correspondences, with a $1$ indicating a correspondence between a pair of points. 
To build the ground-truth assignment matrix $\bM$, we transform $\bX$ using the ground-truth transformation $\bT$, giving us $\btX$. We then compute the pairwise Euclidean distance matrix between $\btX$ and $\bY$, which we threshold to obtain a correspondence matrix $\bM \in \{0, 1\}$. We augment $\bM$ with an extra row and column acting as outlier bins to obtain $\bbM$. The points without any correspondence are treated as outliers, and the corresponding positions in $\bbM$ are set to one. This strategy does not guarantee a bipartite matching, which we address using a forward-backward check.

We then express our loss function as the negative log-likelihood
\begin{equation}
    \mathcal{L}(\bbP, \bbM) = \frac{- \sum\limits_{i =1}^{M}\sum\limits_{j = 1}^{N} (\log \bbP_{i,j})\bbM_{i, j}}{\sum\limits_{i = 1}^{M}\sum\limits_{j = 1}^{N} \bbM_{i, j}}\;,
    \label{eq:nll}
\end{equation}
where $\bbP$ is the estimated score map, and where the denominator normalizes the loss value so that different training samples containing different number of correspondences have the same influence in the overall empirical risk.

\subsection{Network Architectures}

In this section, we present the two architectures in which we implemented our strategies. One of them relies on point-wise features whereas the other first extract keypoints, thus illustrating the generality of our contributions.
\begin{figure}[!t]
    \centering
    \includegraphics[width=.23\textwidth]{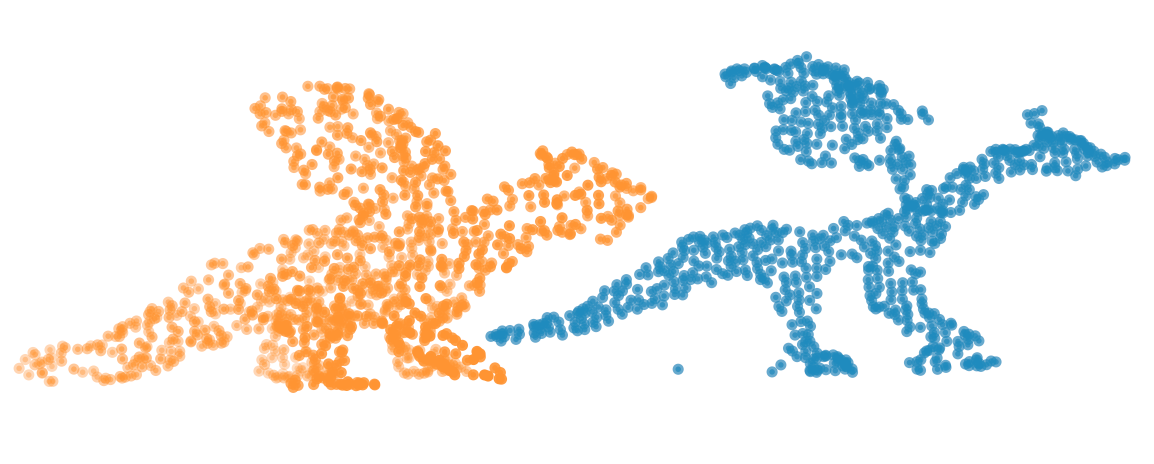}
    \includegraphics[width=.23\textwidth]{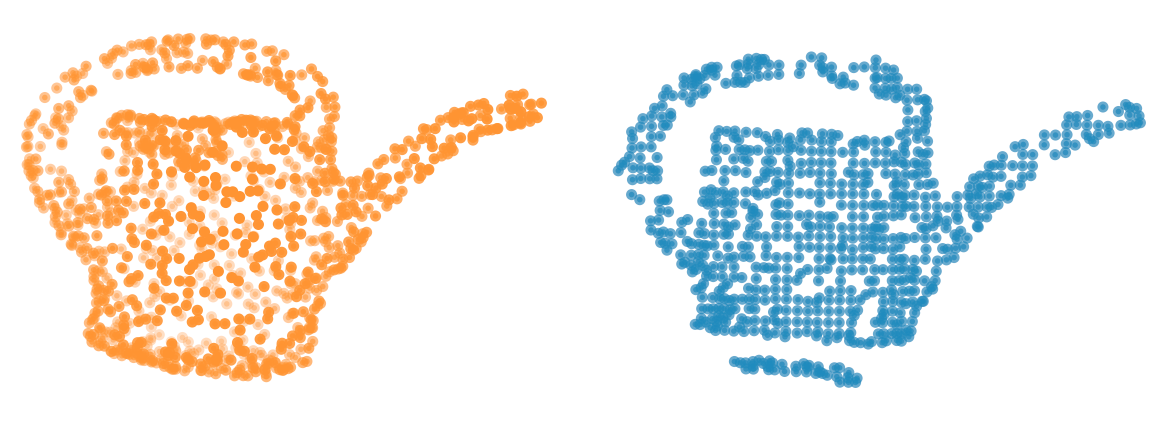}
    \includegraphics[width=.23\textwidth]{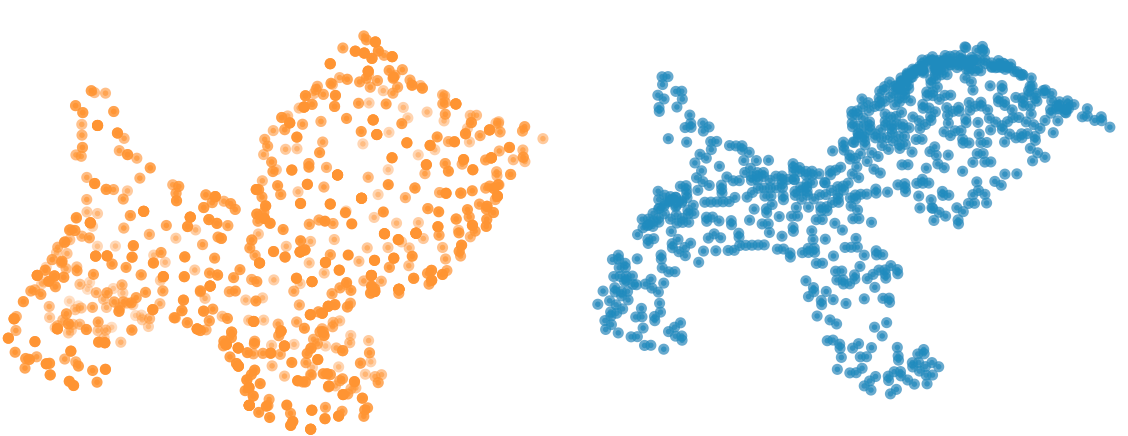}
    \includegraphics[width=.23\textwidth]{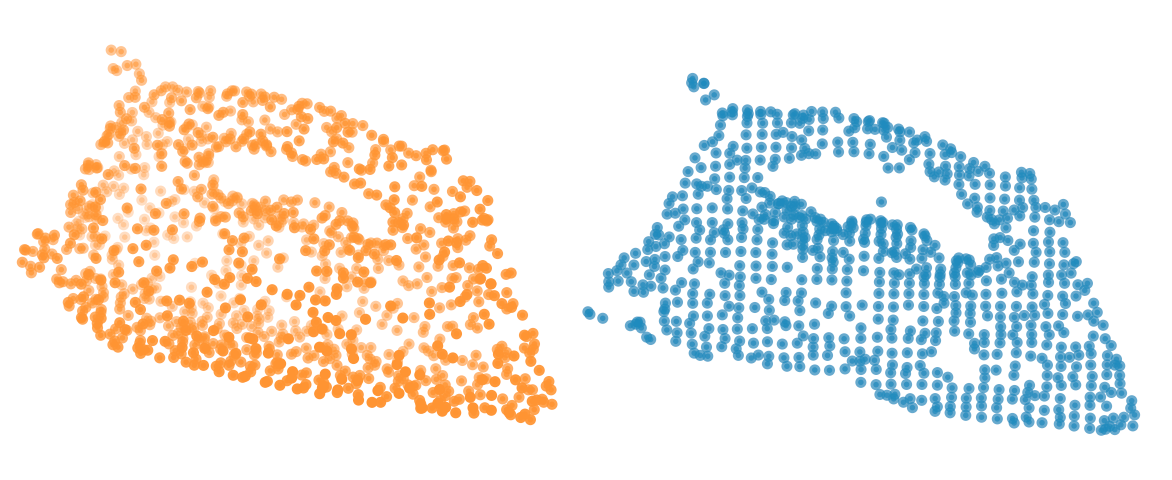}
    \includegraphics[width=.23\textwidth]{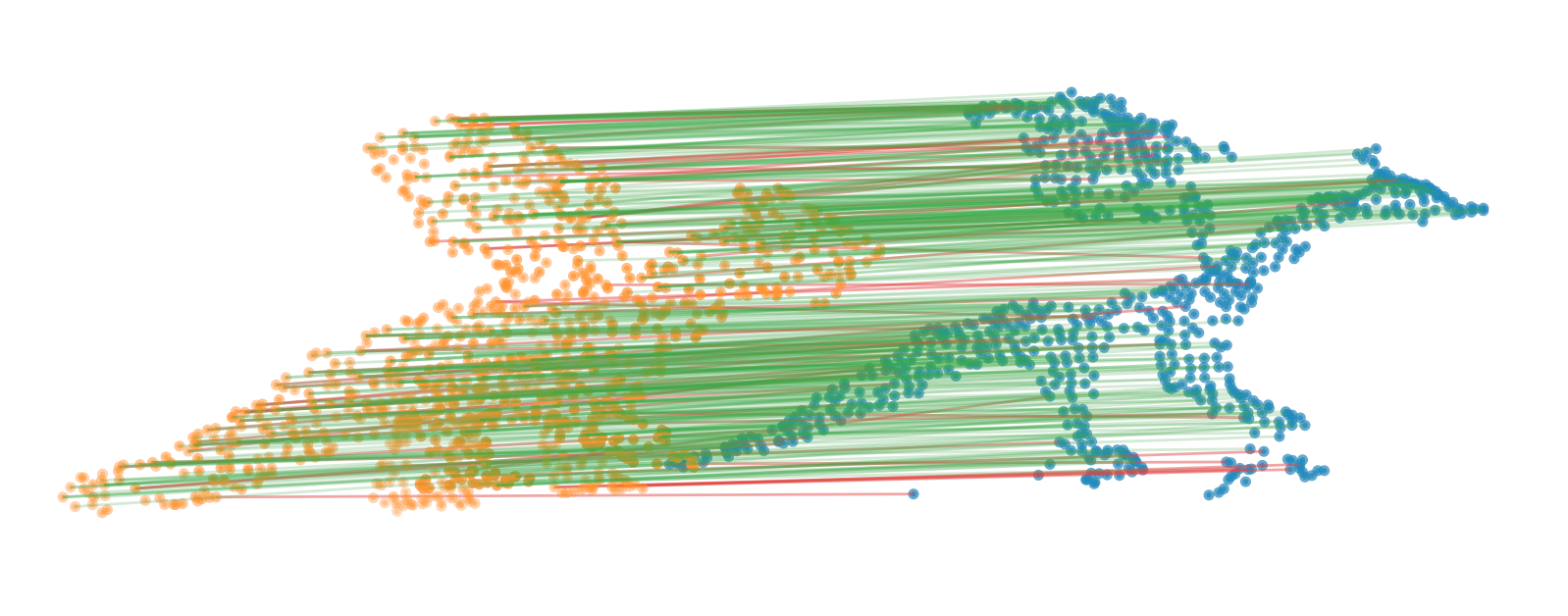}
    \includegraphics[width=.23\textwidth]{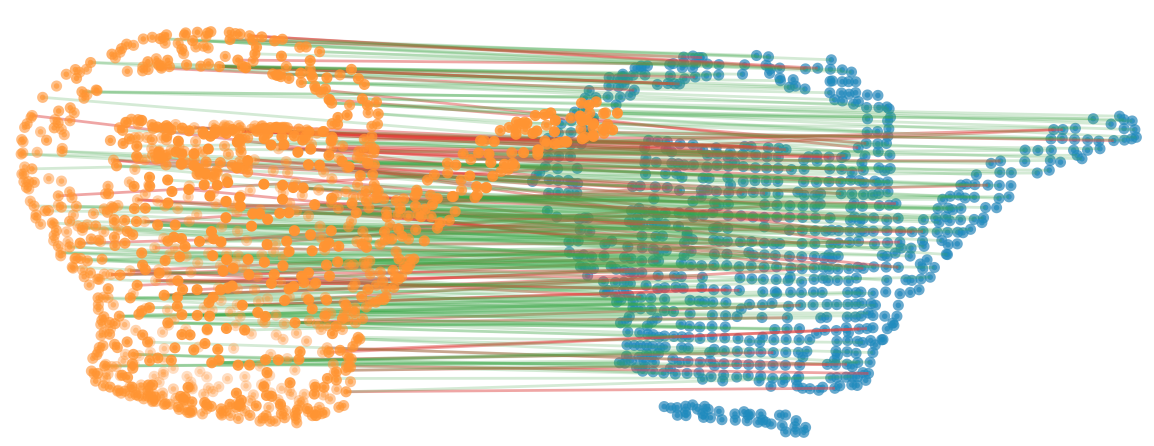}
    \includegraphics[width=.23\textwidth]{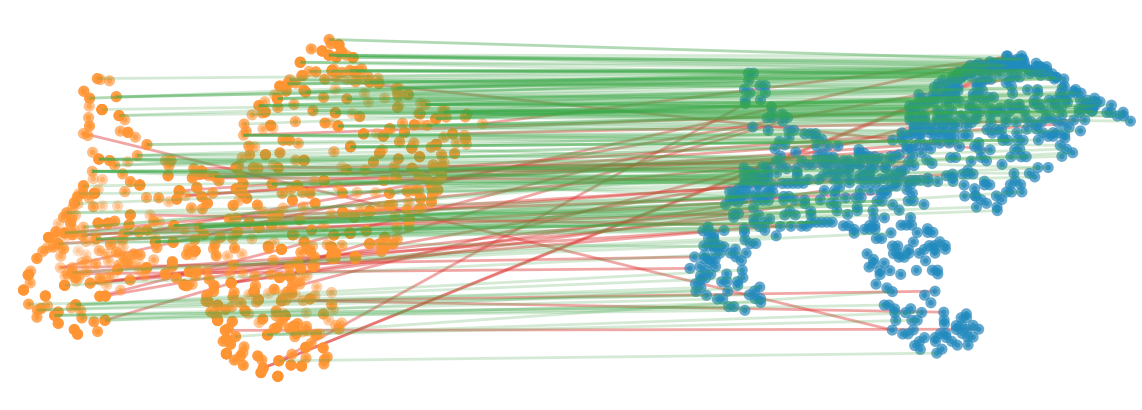}
    \includegraphics[width=.23\textwidth]{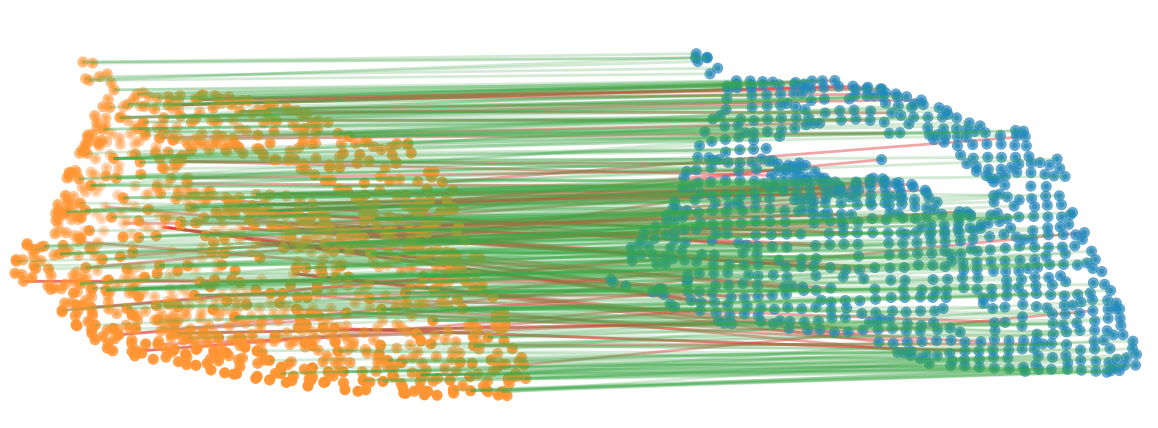}
    \includegraphics[width=.26\textwidth]{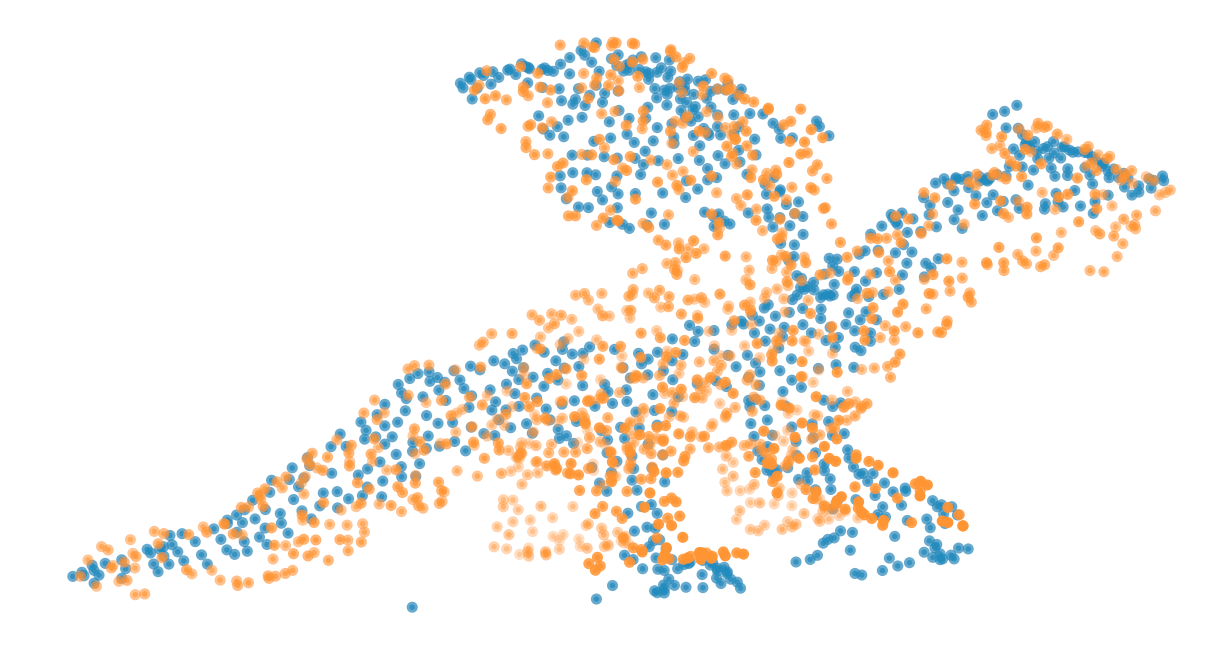}
    \includegraphics[width=.21\textwidth]{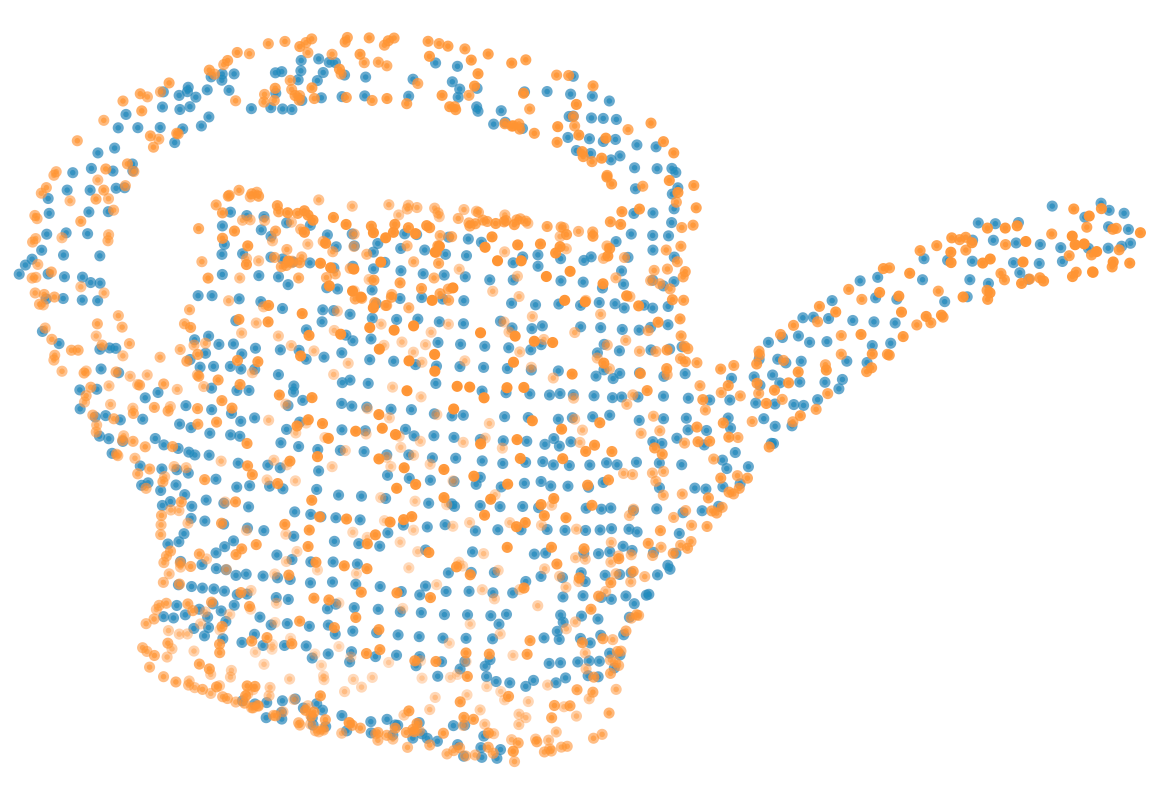}
    \includegraphics[width=.18\textwidth]{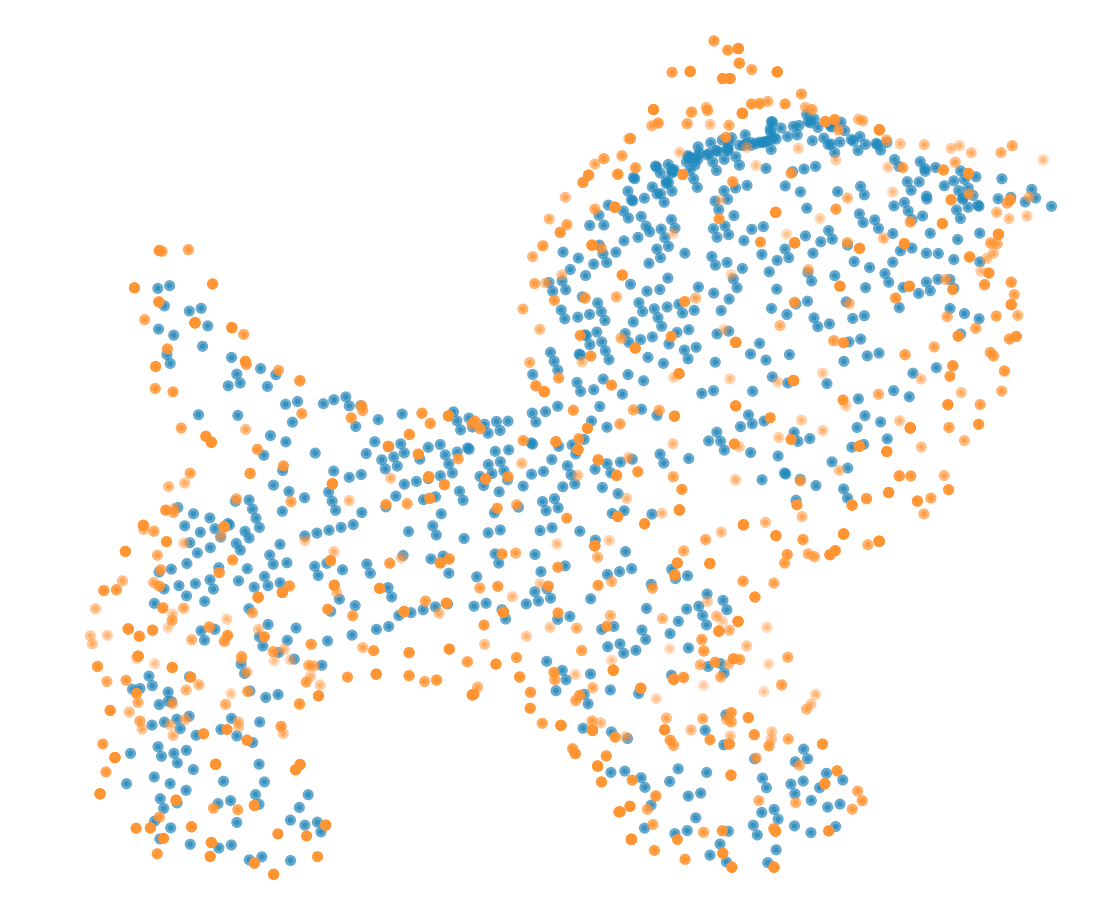}
    \includegraphics[width=.23\textwidth]{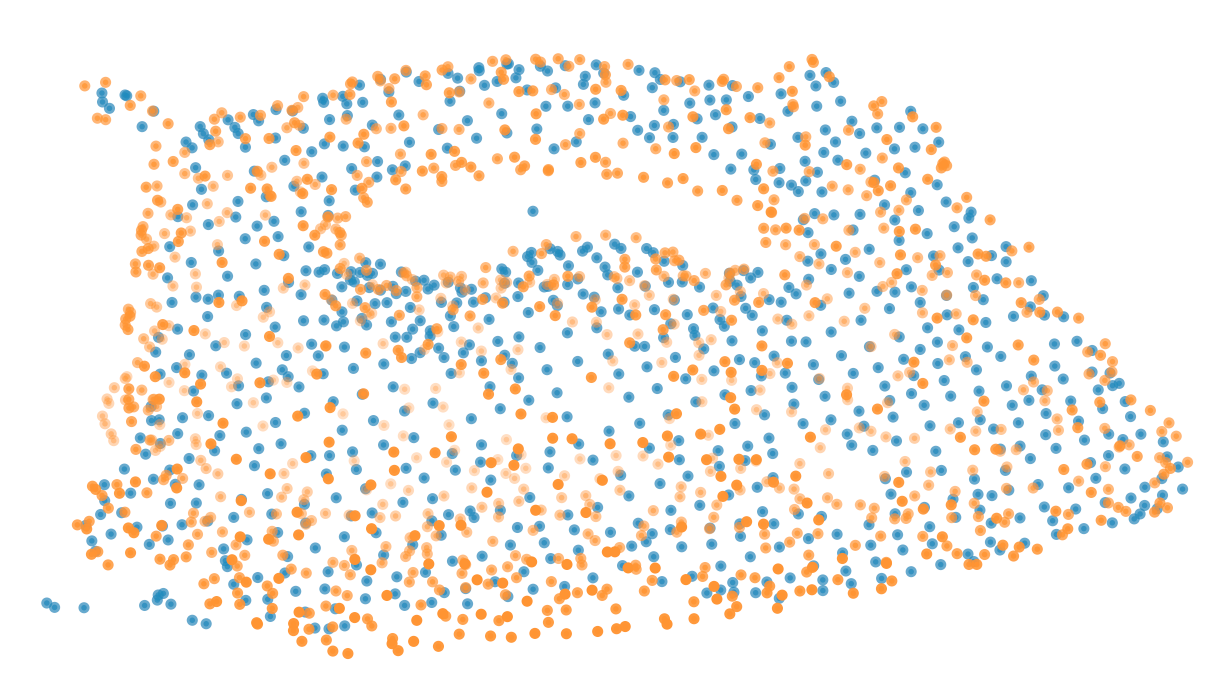}
    \caption{\label{fig:matches} {\bf Qualitative results of Ours-DCP+ICP.} {\bf Top:} Input source (orange) and target (blue) point clouds. We show objects from the TUD-L (dragon,watering can), LINEMOD (kitten) and Occluded-LINEMOD (iron) datasets. {\bf Middle:} Matches found by Ours-DCP+ICP, with the true inlier matches in green, and the outlier matches in red. {\bf Bottom:} Registration results, showing that the source and target sets are correctly aligned.}
\end{figure}

\subsubsection{DCP-based Architecture.}

In DCPv2, the feature extraction module, denoted by $\Theta(\cdot,\cdot)$, takes two point sets as input, and outputs the feature matrix $\bthx$, resp. $\bthy$, i.e., one $P$-dimensional feature vector per 3D point for $\bX$, resp. $\bY$. Then then two feature matrices be passed to a transformer,
which learns a function $\phi : \mathbb{R}^{M\times P} \times \mathbb{R}^{N\times P} \rightarrow \mathbb{R}^{M\times P}$, that combines the information of the two point sets. Ultimately, this produces the descriptor matrix $\bfx$, resp. $\bfy$, for $\bX$, resp. $\bY$, written as
\begin{equation}
    \bfx = \bthx + \phi(\bthx, \bthy), \;\;\;
    \bfy = \bthy + \phi(\bthy, \bthx)\;.
\end{equation}
For our architecture, we integrate our Match Normalization strategy to the layers of the feature extractor $\Theta(\cdot,\cdot)$, while keeping the transformer architecture unchanged.


Inspired by previous work~\cite{Sarlin19,Yew20}, we choose to use a Sinkhorn layer to handle the outliers. Specifically, we extend the score matrix $\bS$ of Eq.~\ref{eq:score} by one row and one column to form an augmented score matrix $\bbS$. The values at the newly-created positions in $\bbS$ are set to
\begin{equation}
    \bbS_{i, N + 1} = \bbS_{M + 1, j} = \bbS_{M + 1, N + 1} = \alpha,
\end{equation}
$\forall i \in [1, M], \;\forall j \in [1, N]$, where $\alpha \in \mathbb{R}$ is a fixed parameter, which we set to $1$ in practice. The values at the other indices directly come from $\bS$. Given the augmented score map $\bbS$, we aim to find a partial assignment $\bbP \in \mathbb{R}^{(M+1)\times (N+1)}$, defining correspondences between the two point sets, extended with the outlier bins. This assignment is obtained by a differentiable version of the Sinkhorn algorithm~\cite{Yew20,Sarlin19}, and is used in the calculation of the loss function of Eq.~\ref{eq:score}. At test time, we use the output of the Sinkhorn layer to ﬁnd the best set of corresponding points between the two point clouds. In addition to the points found as outliers, we also discard those whose value in the score map are below a threshold.

\subsubsection{IDAM-based Architecture.}
The main difference compared to the DCP-based architecture is that this architecture builds the score map upon selected keypoints instead of all the input points. Similarly to the DCP-based architecture, the IDAM-based one uses a feature extraction module $\Theta(\cdot,\cdot)$. This module can be a traditional local descriptor, FPFH~\cite{Rusu09}, or a learning-based method. We therefore integrate our Match Normalization to the learning-based feature extraction network. The extracted features $\bthx$ and $\bthy$ are then passed to a keypoint selection module, corresponding to a second network that outputs a significance score. The significance score is used to obtain a fixed number of keypoints. We denote the features of the reduced keypoint sets as $\bthxt\in\mathbb{R}^{M'\times P}$ and $\bthyt\in\mathbb{R}^{N'\times P}$. 
These features, combined with their corresponding original coordinates, are used to calculate a reduced score map $\bS'\in\mathbb{R}^{M'\times N'\times (2P + 4)}$, which is processed by a similarity matrix convolutional neural network to obtain the final score map  used to find the matches. IDAM further incorporates an iterative registration loop to this process to refine the results, as illustrated in Figure~\ref{fig:arch}.
Please refer to~\cite{Li20} for more detail.

For IDAM, the loss function is composed of three terms: One to supervise the score matrix as in the DCP-based architecture, and two to supervise the keypoint selection network. 
The framework assumes that the outliers have been eliminated in the keypoint selection process. Therefore, at test time, we simply compute the argmax of each row to find the best matches. More details can be found in the paper~\cite{Li20}.

\section{Experiments}
\subsection{Datasets and Training Parameters}
We evaluate our method on three object-level pose estimation real scene datasets: TUD-L~\cite{Hodan18}, LINEMOD~\cite{Hinterstoisser12}, Occluded-LINEMOD~\cite{Brachmann14}. 
For TUD-L, we use the provided real scene training data for training. As there are only 1214 testing images and no explicit training data in Occluded-LINEMOD, we train our network based on the LINEMOD training data. To be specific, we use the PBR dataset provided by the BOP Benchmark~\cite{Hodan18}. For testing, we follow the BOP 2019 challenge instructions and use the provided testing split for testing. 
We implement our DCP-based pose estimation network in Pytorch~\cite{Paszke17} and train it from scratch. We use the Adam optimizer~\cite{Kingma15} with a learning rate of $10^{-3}$ and mini-batches of size $32$, and train the network for $30,000$ iterations.  For the OT layer, we use $k = 50$ iterations and set $\lambda = 0.5$. For our IDAM-based architecture, we train the model with the Adam optimizer~\cite{Kingma15} until convergence, using a learning rate of $10^{-4}$ and mini-batches of size $32$. We use the FPFH implementation from the Open3D~\cite{Zhou18} library and our custom DGCNN for feature extraction. We set the number of refinement iterations for both the FPFH-based and DGCNN-based versions to 3. 
For both frameworks, we set the number of points for $\bX$ and $\bY$ to be $1024$ and $768$, respectively, encoding the fact that $\bY$ only contains a visible portion of $\bX$. To obtain the target point clouds from the depth maps, we use the masks provided with the datasets. Training was performed on one NVIDIA RTX8000 GPU. 

\subsection{Evaluation Metrics}

For evaluation, in addition to the three metrics used by the BOP benchmark, Visible Surface Discrepancy (VSD)~\cite{Hodan16,Hodan18}, Maximum Symmetry-Aware Surface Distance (MSSD)~\cite{Drost17} and Maximum Symmetry-Aware Projection Distance (MSPD)~\cite{Brachmann16}, we also report the rotation and translation error between the predictions $\hR,\hht$ and the ground truth $\gR,\gt$. These errors are computed as
\begin{equation}
    \begin{aligned}
        E_{rot}(\hR,\gR) = arccos\frac{trace(\hR^{\top}\gR) - 1}{2} \;,
        E_{trans}(\hht,\gt) = \norm{\hht - \gt}^{2}_2\;.
    \end{aligned}
    \label{syn_metric}
\end{equation}
We summarize the results in terms of mean average precision (mAP) of the estimated relative pose under varying accuracy thresholds, as in~\cite{Dang18}.
We keep the rotation error unchanged. Furthermore, we also report the ADD metric~\cite{Xiang18}, which measures the average distance between the 3D model points transformed using the predicted pose and those obtained with the ground-truth one. We set the threshold to be $10\%$ of the model diameter, as commonly done in 6D pose estimation.

\begin{figure*}[!t]
    \centering
    \begin{subfigure}{.24\textwidth}{\centering\includegraphics[width=\linewidth,trim=150 50 0 150,clip]{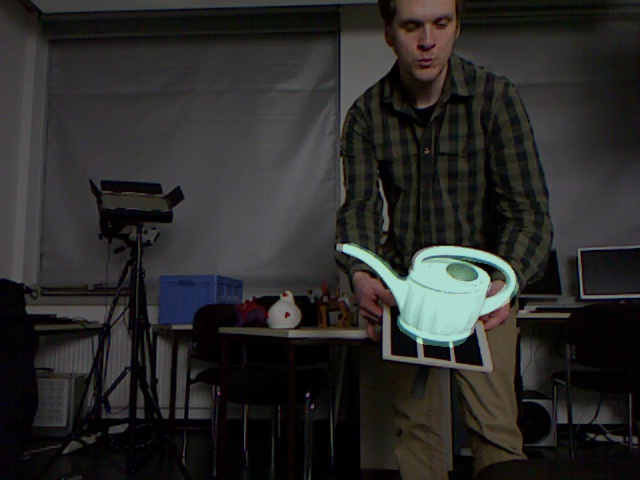}}\end{subfigure}
    \begin{subfigure}{.24\textwidth}{\centering\includegraphics[width=\linewidth,trim=150 50 0 150,clip]{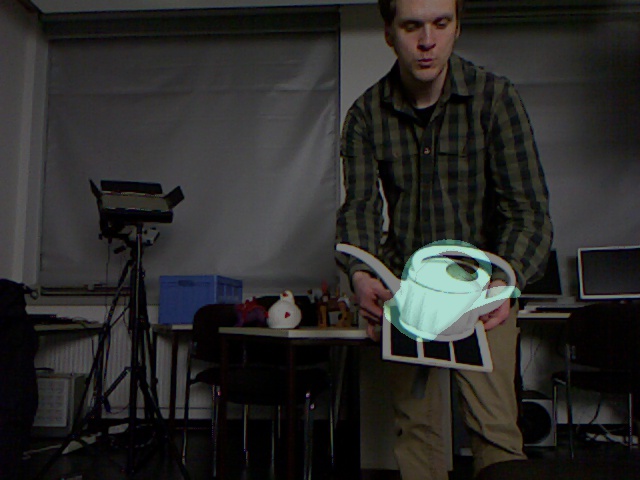}}\end{subfigure}
    \begin{subfigure}{.24\textwidth}{\centering\includegraphics[width=\linewidth,trim=150 50 0 150,clip]{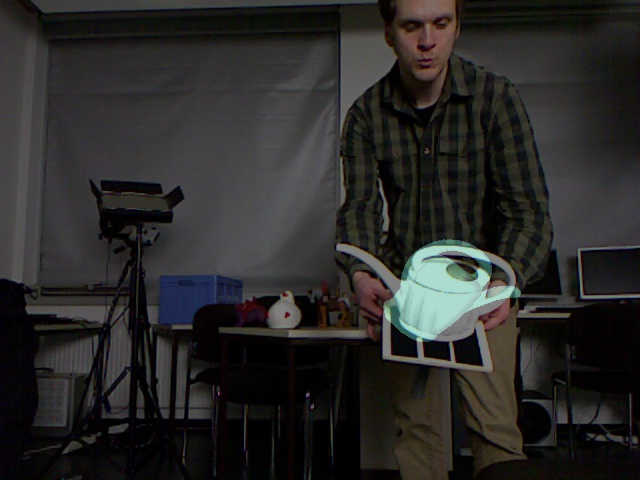}}\end{subfigure}
    \begin{subfigure}{.24\textwidth}{\centering\includegraphics[width=\linewidth,trim=150 50 0 150,clip]{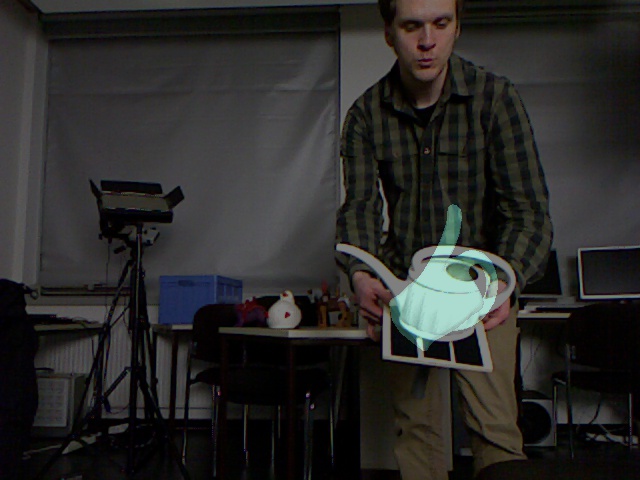}}\end{subfigure}
    
    \begin{subfigure}{.24\textwidth}{\centering\includegraphics[width=\linewidth,trim=100 100 50 100,clip]{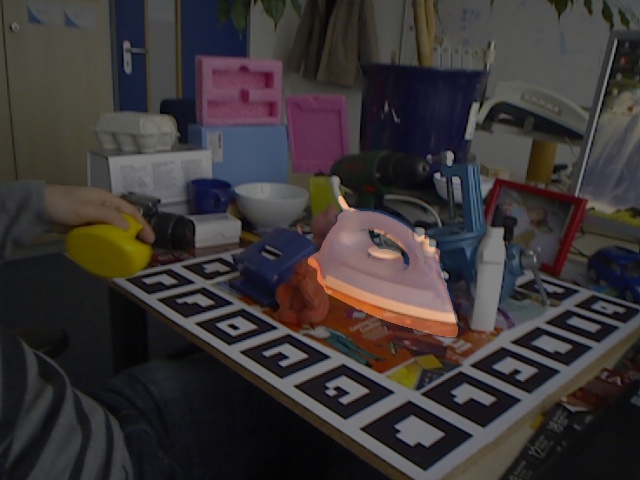}}\end{subfigure}
    \begin{subfigure}{.24\textwidth}{\centering\includegraphics[width=\linewidth,trim=100 100 50 100,clip]{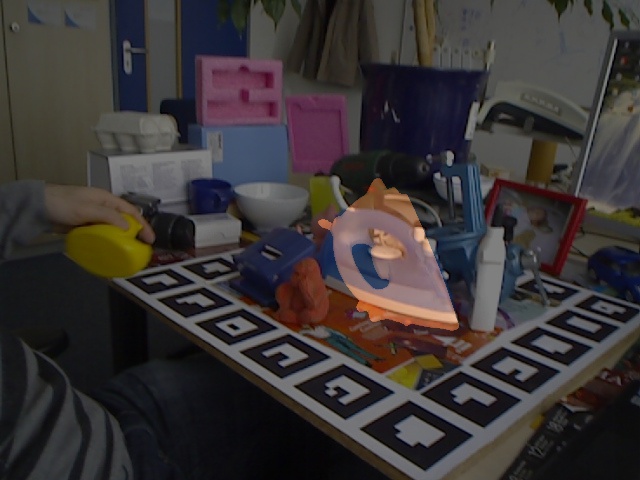}}\end{subfigure}
    \begin{subfigure}{.24\textwidth}{\centering\includegraphics[width=\linewidth,trim=100 100 50 100,clip]{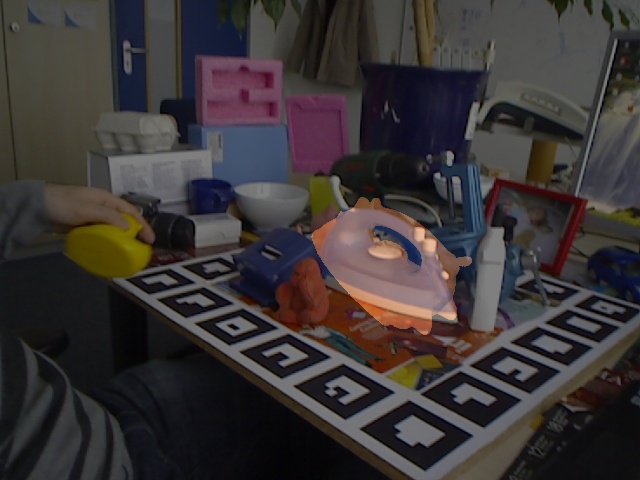}}\end{subfigure}
    \begin{subfigure}{.24\textwidth}{\centering\includegraphics[width=\linewidth,trim=100 100 50 100,clip]{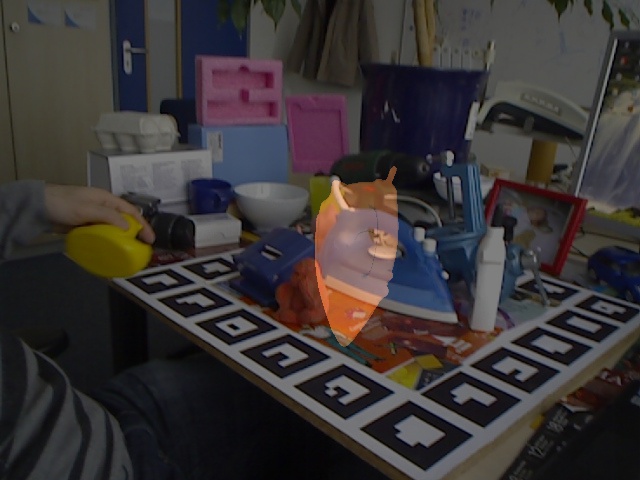}}\end{subfigure}

    \begin{subfigure}{.24\textwidth}{\centering\includegraphics[width=\linewidth,trim=0 150 150 50,clip]{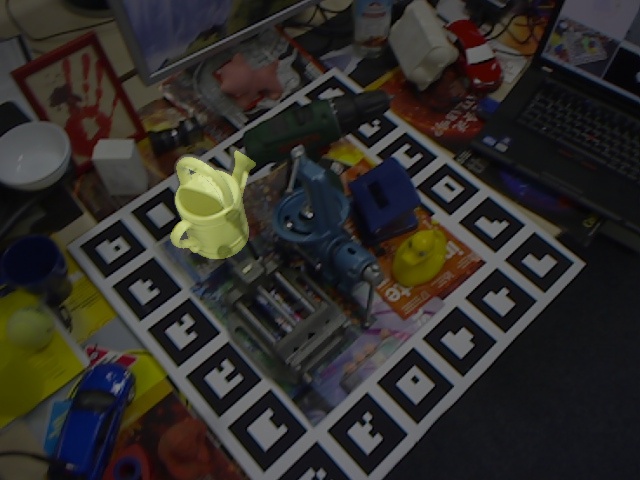}\caption{Ours-DCP+ICP}}\end{subfigure}
    \begin{subfigure}{.24\textwidth}{\centering\includegraphics[width=\linewidth,trim=0 150 150 50,clip]{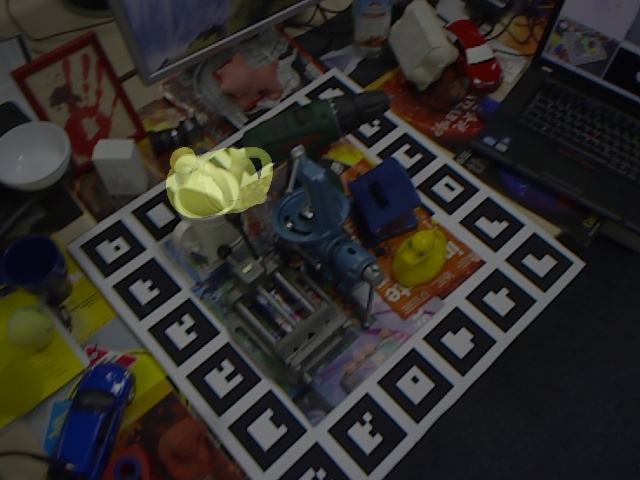}\caption{Ous-IDAM+ICP}}\end{subfigure}
    \begin{subfigure}{.24\textwidth}{\centering\includegraphics[width=\linewidth,trim=0 150 150 50,clip]{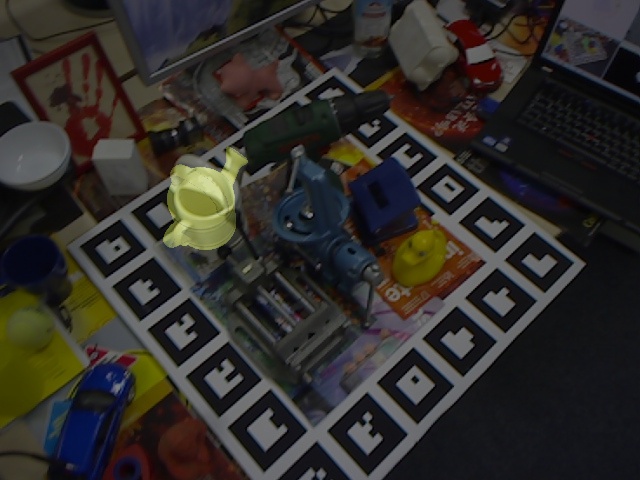}\caption{Super4PCS}}\end{subfigure}
    \begin{subfigure}{.24\textwidth}{\centering\includegraphics[width=\linewidth,trim=0 150 150 50,clip]{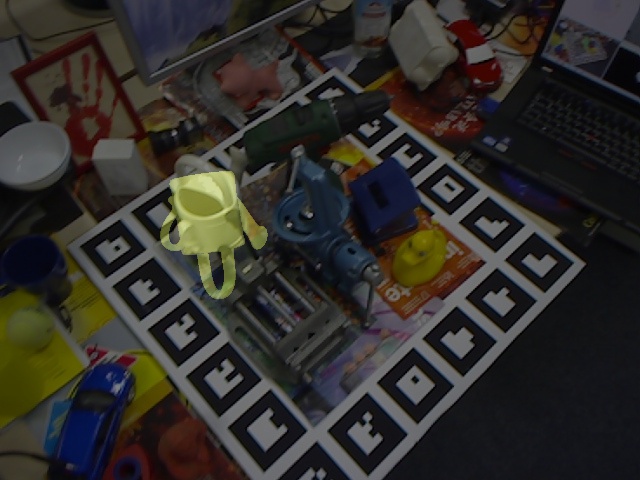}\caption{Teaser++}}\end{subfigure}
    
    \caption{\label{visualization} Qualitative results on the TUD-L (top), LINEMOD (middle) and Occluded-LINEMOD (bottom) datasets.}

\end{figure*}

\subsection{Comparison with Existing Methods}

We compare our method to both traditional techniques and learning-based ones.
Specifically, for the traditional methods, we used the Open3D~\cite{Zhou18} implementations of ICP~\cite{Besl92a} and FGR~\cite{Zhou16}, the official implementation of \href{https://github.com/MIT-SPARK/TEASER-plusplus}{TEASER++}~\cite{Yang19}, and the author's binary file for \href{https://github.com/nmellado/Super4PCS}{Super4PCS}~\cite{Mellado14}. For DCP, we used the default DCPv2 training settings.
However, because of instabilities caused by the SVD computation, we had to train the model several times to eventually find a point before a crash achieving reasonable accuracy. Note that this problem was also reported in~\cite{Wang19g}.
We also tried to train PRNet and RPMNet, but failed to get reasonable results because of similar SVD-related crashes, as also observed in~\cite{Choy20}; specifically, SVD always crashes at the beginning of training when processing real data.
For IDAM, we report results using both traditional FPFH features and features extracted with a DGCNN.
Ours-DCP denotes our approach implemented in the DCPv2 architecture, where we replace the SVD-based loss with the NLL one, replace the softmax layer with a Sinkhorn layer, and integrate Match Normalization with the DGCNN. Ours-IDAM denotes our approach within the IDAM architecture, incorporating Match Normalization in the DGCNN. Since the IDAM does not use an SVD-based loss function, we retained its original loss.

\subsubsection{TUD-L dataset.}
\begin{table*}[!t]
    \centering
    {\scriptsize
    \setlength{\tabcolsep}{1.mm}{
    \begin{tabular}{l|ccc|ccc|c|cccc}
        \toprule
        & \multicolumn{3}{c|}{Rotation mAP} & \multicolumn{3}{c|}{Translation mAP} & \multicolumn{1}{c|}{ADD} & \multicolumn{4}{c}{BOP Benchmark} \\
        Method & $5^{\circ}$ & $10^{\circ}$ & $20^{\circ}$ & $1cm$ & $2cm$ & $5cm$ & $0.1d$ & VSD & MSSD & MSPD & AR\\
        \midrule
        ICP                 & 0.02 & 0.02 & 0.02 & 0.01 & 0.14 & 0.57 & 0.02 & 0.117 & 0.023 & 0.027 & 0.056 \\
        \rowcolor{tabgray}
        FGR(FPFH)           & 0.00 & 0.01 & 0.01 & 0.04 & 0.25 & 0.63 & 0.01 & 0.071 & 0.007 & 0.008 & 0.029 \\
        TEASER++(FPFH)      & 0.13 & 0.17 & 0.19 & 0.03 & 0.22 & 0.56 & 0.17 & 0.175 & 0.196 & 0.193 & 0.188 \\
        \rowcolor{tabgray}
        Super4PCS           & 0.30 & 0.50 & 0.56 & 0.05 & 0.40 & 0.92 & 0.54 & 0.265 & 0.500 & 0.488 & 0.418 \\
        DCP(v2)             & 0.00 & 0.01 & 0.02 & 0.02 & 0.07 & 0.55 & 0.01 & 0.0253 & 0.051 & 0.039 & 0.038 \\ 
        \rowcolor{tabgray}
        IDAM(FPFH)          & 0.05 & 0.12 & 0.20 & 0.03 & 0.17 & 0.63 & 0.13 & 0.100 & 0.194 & 0.166 & 0.153 \\
        IDAM                & 0.03 & 0.05 & 0.10 & 0.02 & 0.08 & 0.49 & 0.05 & 0.067 & 0.108 & 0.099 & 0.091\\ 
        \rowcolor{tabgray}
        $\star$Vidal-Sensors18 & - & - & - & - & - & - & - & 0.811 & 0.910 & 0.907 & 0.876 \\
        $\star$Drost & - & - & - & - & - & - & - & 0.809 & 0.875 & 0.872 & 0.852 \\
        \rowcolor{tabgray}
        Ours-IDAM           & 0.36 & 0.46 & 0.53 & 0.23 & 0.47 & 0.75 & 0.46 & 0.339 & 0.502 & 0.492 & 0.444\\ 
        Ours-IDAM+ICP       & 0.56 & 0.58 & 0.61 & 0.55 & 0.66 & 0.81 & 0.58 & 0.580 & 0.604 & 0.618 & 0.601\\ 
        \rowcolor{tabgray}
        Ours-DCP            & 0.70 & 0.81 & 0.87 & 0.71 & 0.86 & 0.97 & 0.85 & 0.700 & 0.853 & 0.852 & 0.801\\
        Ours-DCP+ICP       & \textbf{0.91}& \textbf{0.92}& \textbf{0.93}& \textbf{0.86}& \textbf{0.95}& \textbf{0.99}& \textbf{0.93}& \textbf{0.859} & \textbf{0.914} & \textbf{0.935} & \textbf{0.903}\\
        \bottomrule
    \end{tabular}
    }}
    \caption{Quantitative comparison of our method with previous work on the \textbf{TUD-L} real scene dataset. The results for Vidal-Sensor18~\cite{Vidal18} and Drost (Drost-CVPR10-3D-Edges)~\cite{Drost10} were directly taken from the BOP leaderboard, and, in contrast with all the other results, were obtained without using a mask for the target point cloud. 
    Our contributions allow existing learning-based methods, such as IDAM and DCP, to successfully register real-world data.}
    \label{tab:tudl_comb}

\end{table*}

The results of all methods for the TUD-L dataset are summarized in Table~\ref{tab:tudl_comb}. Note that the traditional methods based on FPFH features yield poor results, because FPFH yields unreliable features in the presence of many smooth areas on the objects. Vanilla DCPv2 and IDAM also struggle with such real-world data. However, these baselines are significantly improved by our Match Normalization strategy, Ours-DCP and Ours-IDAM, both of which outperform Super4PCS. Our results can further be boosted by the use of ICP as a post-processing step.


In the table, we also provide the results of `Vidal-Sensors18' and `Drost-CVPR10-3D-Edges', the two best depth-only performers in the BOP leaderboard. Note that these are traditional methods whose results were obtained without using a mask to segment the target point cloud, which makes the comparison favorable to our approach. Nevertheless, our results evidence that our contributions can make learning-based 3D object registration applicable to real-world data, which we believe to be a significant progress in the field.

\subsubsection{LINEMOD dataset.}
\begin{table*}[!t]
    \centering
    {\scriptsize
    \setlength{\tabcolsep}{1mm}{
    \begin{tabular}{l|ccc|ccc|c|cccc}
        \toprule
        & \multicolumn{3}{c|}{Rotation mAP} & \multicolumn{3}{c|}{Translation mAP} & \multicolumn{1}{c|}{ADD} & \multicolumn{4}{c}{BOP Benchmark} \\
        Method & $5^{\circ}$ & $10^{\circ}$ & $20^{\circ}$ & $1cm$ & $2cm$ & $5cm$ & $0.1d$ & VSD & MSSD & MSPD & AR\\
        \midrule
        ICP                 & 0.00 & 0.01 & 0.01 & 0.04 & 0.27 & 0.82 & 0.01 & 0.092 & 0.014 & 0.027 & 0.044 \\
        \rowcolor{tabgray}
        FGR(FPFH)           & 0.00 & 0.00 & 0.00 & 0.05 & 0.31 & 0.89 & 0.00 & 0.068 & 0.000 & 0.010 & 0.026 \\
        TEASER++(FPFH)      & 0.01 & 0.03 & 0.05 & 0.03 & 0.21 & 0.73 & 0.03 & 0.108 & 0.076 & 0.098 & 0.094 \\
        \rowcolor{tabgray}
        Super4PCS           & 0.02 & 0.09 & 0.15 & 0.04 & 0.31 & 0.89 & 0.10 & 0.117 & 0.178 & 0.201 & 0.165 \\
        DCP(v2)             & 0.00 & 0.00 & 0.01 & 0.05 & 0.24 & 0.83 & 0.00 & 0.057 & 0.025 & 0.049 & 0.044 \\ 
        \rowcolor{tabgray}
        IDAM(FPFH)          & 0.00 & 0.01 & 0.03 & 0.03 & 0.16 & 0.67 & 0.01 & 0.053 & 0.055 & 0.069 & 0.059 \\
        IDAM                & 0.00 & 0.01 & 0.05 & 0.03 & 0.16 & 0.71 & 0.02 & 0.050 & 0.070 & 0.081 & 0.067 \\ 
        \rowcolor{tabgray}
        $\star$PPF\_3D\_ICP & - & - & - & - & - & - & - & \textbf{0.719} & \textbf{0.856} & \textbf{0.866} & \textbf{0.814} \\ 
        $\star$Drost & - & - & - & - & - & - & - & 0.678 & 0.786 & 0.789 & 0.751 \\
        \rowcolor{tabgray}
        Ours-IDAM            & 0.01 & 0.07 & 0.15 & 0.13 & 0.38 & 0.87 & 0.11 & 0.148 & 0.194 & 0.209 & 0.184 \\ 
        Ours-IDAM+ICP        & 0.15 & 0.23 & 0.27 & 0.25 & 0.54 & 0.91 & 0.23 & 0.352 & 0.311 & 0.345 & 0.336 \\ 
        \rowcolor{tabgray}
        Ours-DCP            & 0.10 & 0.27 & 0.49 & 0.26 & 0.60 & 0.95 & 0.37 & 0.319 & 0.490 & 0.529 & 0.446 \\
        Ours-DCP+ICP        & \textbf{0.43} & \textbf{0.59} & \textbf{0.67} & \textbf{0.49} & \textbf{0.83} & \textbf{0.97} & \textbf{0.60} & 0.616 & 0.680 & 0.737 & 0.678 \\
        \bottomrule
    \end{tabular}
    }}
    \caption{Quantitative comparison of our method with previous work on the \textbf{LINEMOD} real scene dataset. PPF\_3D\_ICP~\cite{Drost10} and Drost (Drost-CVPR10-3D-Only)~\cite{Drost10} are traditional methods and represent the best depth-only performers from the BOP leaderboard.}
    \label{tab:lm_comb}
\end{table*}

In contrast with TUD-L, the LINEMOD dataset contains symmetrical objects and small occlusions at the object boundaries, which increase the difficulty of this dataset.
As shown in Table~\ref{tab:lm_comb}, even Super4PCS is therefore unable to yield meaningful results on this dataset.
Furthermore, as the LINEMOD training data does not contain any real-world measurements, the training-testing domain gap further complicates the task for learning-based methods. Nevertheless, our approach improves the results of both DCP and IDAM, allowing them to produce reasonably accurate pose estimates.


\subsubsection{Occluded-LINEMOD dataset.}
\begin{table*}[!t]
    \centering
    {\scriptsize
    \setlength{\tabcolsep}{1mm}{
    \begin{tabular}{l|ccc|ccc|c|cccc}
        \toprule
        & \multicolumn{3}{c|}{Rotation mAP} & \multicolumn{3}{c|}{Translation mAP} & \multicolumn{1}{c|}{ADD} & \multicolumn{4}{c}{BOP Benchmark} \\
        Method & $5^{\circ}$ & $10^{\circ}$ & $20^{\circ}$ & $1cm$ & $2cm$ & $5cm$ & $0.1d$ & VSD & MSSD & MSPD & AR\\
        \midrule
        ICP                 & 0.01 & 0.01 & 0.01 & 0.07 & 0.36 & 0.85 & 0.01 & 0.085 & 0.014 & 0.032 & 0.044  \\
        \rowcolor{tabgray}
        FGR(FPFH)           & 0.00 & 0.00 & 0.00 & 0.08 & 0.43 & 0.85 & 0.00 & 0.055 & 0.000 & 0.009 & 0.021  \\
        TEASER++(FPFH)      & 0.01 & 0.02 & 0.05 & 0.04 & 0.26 & 0.77 & 0.02 & 0.096 & 0.060 & 0.093 & 0.083  \\
        \rowcolor{tabgray}
        Super4PCS           & 0.01 & 0.03 & 0.06 & 0.06 & 0.31 & 0.83 & 0.03 & 0.054 & 0.072 & 0.113 & 0.080 \\
        DCP(v2)             & 0.00 & 0.00 & 0.01 & 0.03 & 0.30 & 0.83 & 0.00 & 0.055 & 0.018 & 0.059 & 0.044 \\ 
        \rowcolor{tabgray}
        IDAM(FPFH)          & 0.00 & 0.00 & 0.02 & 0.04 & 0.18 & 0.73 & 0.00 & 0.044 & 0.033 & 0.066 & 0.048 \\
        IDAM                & 0.00 & 0.02 & 0.06 & 0.07 & 0.26 & 0.76 & 0.02 & 0.063 & 0.088 & 0.119 & 0.090 \\ 
        \rowcolor{tabgray}
        $\star$Vidal-Sensors18 & - & - & - & - & - & - & - & 0.473 & 0.625 & 0.647 & 0.582 \\
        $\star$PPF\_3D\_ICP & - & - & - & - & - & - & - & \textbf{0.523} & \textbf{0.669} & \textbf{0.716} & \textbf{0.636} \\
        \rowcolor{tabgray}
        Ours-IDAM            & 0.02 & 0.08 & 0.18 & 0.15 & 0.44 & 0.84 & 0.12 & 0.155 & 0.204 & 0.248 & 0.202 \\ 
        Ours-IDAM+ICP        & 0.15 & 0.22 & 0.32 & 0.23 & 0.58 & 0.88 & 0.25 & 0.349 & 0.320 & 0.374 & 0.348 \\ 
        \rowcolor{tabgray}
        Ours-DCP                & 0.07 & 0.19 & 0.36 & 0.24 & 0.57 & 0.88 & 0.28 & 0.263 & 0.384 & 0.450 & 0.365 \\
        Ours-DCP+ICP          & \textbf{0.31} & \textbf{0.46} & \textbf{0.56} & \textbf{0.37} & \textbf{0.70} & \textbf{0.91} & \textbf{0.47} & 0.478 & 0.542 & 0.612 & 0.544 \\
        \bottomrule
    \end{tabular}
    }}
    \caption{Quantitative comparison of our method with previous work on the \textbf{Occluded-LINEMOD} real scene dataset. Vidal-Sensors18~\cite{Vidal18} and PPF\_3D\_ICP~\cite{Drost10} are traditional methods and represent the best depth-only performers from the BOP leaderboard.}
    \label{tab:lmo_comb}
\end{table*}

The Occluded-LINEMOD dataset further increases the challenge compared to LINEMOD by adding severe occlusions, in addition to the still-existing domain gap. As such, as shown in Table~\ref{tab:lmo_comb}, the results of all the methods deteriorate.  Nevertheless, our approach still allows DCP and IDAM to yield meaningful pose estimates.


\subsection{Ablation Study}
In this section, we conduct an ablation study to justify the effectiveness of the proposed Match Normalization. Specifically, 
we evaluate our proposed DCP-based architecture with and without Match Normalization on the TUD-L dataset.
In addition to the metrics used in the previous section, we report the number of matches predicted by the network and the number of real inliers within these predicted matches. The predicted matches are those extracted directly from the predicted score map $\bbP$.
We set the threshold to identify the true inliers to $0.02$.
\begin{table*}[!t]
    \centering
    {\scriptsize
    \setlength{\tabcolsep}{.95mm}{
    \begin{tabular}{l|ccc|ccc|c|cccc|cc}
        \toprule
        & \multicolumn{3}{c|}{Rotation mAP} & \multicolumn{3}{c|}{Translation mAP} & \multicolumn{1}{c|}{ADD} & \multicolumn{4}{c|}{BOP Benchmark} & \multicolumn{2}{c}{Matches}\\
        Method & $5^{\circ}$ & $10^{\circ}$ & $20^{\circ}$ & $1cm$ & $2cm$ & $5cm$ & $0.1d$ & \tiny{VSD} & \tiny{MSSD} & \tiny{MSPD} & \tiny{AR} & pred & true\\
        \midrule
        Ours w/o MN & 0.21 & 0.22 & 0.24 & 0.22 & 0.32 & 0.61 & 0.23 & 0.27 & 0.27 & 0.27 & 0.27 & 24.84 & 10.44 \\
        Ours        & 0.91 & 0.92 & 0.93 & 0.86 & 0.95 & 0.99 & 0.93 & 0.86 & 0.91 & 0.94 & 0.90 & 276.10 & 262.96 \\
        
        \bottomrule
    \end{tabular}
    }}
    \caption{{\bf Ablation Study.} We evaluate the influence of our Match Normalization strategy in our DCP-based baseline not only on the same metrics as before but also on the number of matches found by the network.}
    \label{tab:ablation}
\end{table*}

As shown in Table~\ref{tab:ablation}, the number of matches significantly increases with our Match Normalization, and a vast majority of them are true inliers. This evidences that normalizing the source and target point sets with Match Normalization indeed helps the network to find correct matches, and eventually improves the pose estimation performance.
Qualitative results obtained with our method are shown in Figure~\ref{fig:matches}.
Importantly, Match Normalization (MN) does not affect the training efficiency as it adds no new learnable parameters.




\section{Conclusion}

We have identified two factors that prevent the existing learning-based 3D object registration methods from working on real-world data. One is the gap between the feature distributions of the source and target point sets. The larger the difference, the fewer inlier matches are found, which deteriorates the performance. Another is the instability in gradient computations when using an SVD-based loss function, which complicates the network's convergence when the data can undergo a full rotation range.
To tackle the first issue, we have proposed a new normalization method, Match Normalization, which encourages the two point sets to have similar feature distributions by sharing the same scaling parameter. For the second problem, we have replaced the SVD-based loss function with a simple yet robust NLL loss function that imposes direct supervision on the score map. Our two solutions are simple, effective and can be easily applied to many existing learning-based 3D registration frameworks. We have integrated them into a DCP-based and an IDAM-based architecture, and have proven the effectiveness of our method on three real-world 6D object pose estimation datasets, TUD-L, LINEMOD and Occluded-LINEMOD. To the best of our knowledge, this is the first time that a learning-based 3D object registration method achieves meaningful results on real-world data. 
\section{Acknowledgements}
Zheng Dang would like to thank to H. Chen for the highly-valuable discussions and for her encouragement. This work was funded in part by the Swiss Innovation Agency (Innosuisse).
%
%
\bibliographystyle{splncs04}
\bibliography{bibtex/string,bibtex/vision}
\end{document}